\documentclass[letterpaper, 10 pt, conference]{ieeeconf}
\IEEEoverridecommandlockouts
\usepackage{cite}
\usepackage{amsmath,amssymb,amsfonts}
\usepackage{algorithmic}
\usepackage[colorinlistoftodos]{todonotes}
\usepackage{graphicx}
\usepackage{textcomp}
\usepackage{xcolor}
\usepackage{soul}
\usepackage{url}

\usepackage{array}
\usepackage{booktabs}

\newboolean{showcomments}
\setboolean{showcomments}{true}

\def\BibTeX{{\rm B\kern-.05em{\sc i\kern-.025em b}\kern-.08em
    T\kern-.1667em\lower.7ex\hbox{E}\kern-.125emX}}
\begin{document}

\title{STL: Surprisingly Tricky Logic (for System Validation)}

% \author{\IEEEauthorblockN{1\textsuperscript{st} Ho Chit Siu}
% \IEEEauthorblockA{\textit{dept. name of organization (of Aff.)} \\
% \textit{name of organization (of Aff.)}\\
% City, Country \\
% email address or ORCID}
% \and
% \IEEEauthorblockN{2\textsuperscript{nd} Given Name Surname}
% \IEEEauthorblockA{\textit{dept. name of organization (of Aff.)} \\
% \textit{name of organization (of Aff.)}\\
% City, Country \\
% email address or ORCID}
% \and
% \IEEEauthorblockN{3\textsuperscript{rd} Given Name Surname}
% \IEEEauthorblockA{\textit{dept. name of organization (of Aff.)} \\
% \textit{name of organization (of Aff.)}\\
% City, Country \\
% email address or ORCID}
% }

\author{Ho Chit Siu, Kevin Leahy, and Makai Mann% <-this % stops a space
\thanks{DISTRIBUTION STATEMENT A. Approved for public release. Distribution is unlimited.}
\thanks{This material is based upon work supported by the Under Secretary of Defense for Research and Engineering under Air Force Contract No. FA8702-15-D-0001. Any opinions, findings, conclusions or recommendations expressed in this material are those of the author(s) and do not necessarily reflect the views of the Under Secretary of Defense for Research and Engineering.}% <-this % stops a space
\thanks{All authors are with Lincoln Laboratory,
Massachusetts Institute of Technology,
Lexington, MA, USA.
        {\tt\small \{hochit.siu, kevin.leahy, makai.mann\}@ll.mit.edu}}
}

% \author{\IEEEauthorblockN{Ho Chit Siu\IEEEauthorrefmark{1},
% Kevin Leahy\IEEEauthorrefmark{2}, and Makai Mann\IEEEauthorrefmark{3}}
% \IEEEauthorblockA{Lincoln Laboratory,
% Massachusetts Institute of Technology\\
% Lexington, MA, USA\\
% Email: \IEEEauthorrefmark{1}HoChit.Siu@ll.mit.edu,
% \IEEEauthorrefmark{2}Kevin.Leahy@ll.mit.edu,
% \IEEEauthorrefmark{3}Makai.Mann@ll.mit.edu}}

\maketitle

\begin{abstract}
Much of the recent work developing formal methods techniques to specify or learn the behavior of autonomous systems is predicated on a belief that formal specifications are interpretable and useful for humans when checking systems. Though frequently asserted, this assumption is rarely tested. We performed a human experiment (N = 62) with a mix of people who were and were not familiar with formal methods beforehand, asking them to validate whether a set of signal temporal logic (STL) constraints would keep an agent out of harm and allow it to complete a task in a gridworld capture-the-flag setting. Validation accuracy was $45\% \pm 20\%$ (mean $\pm$ standard deviation). The ground-truth validity of a specification, subjects' familiarity with formal methods, and subjects' level of education were found to be significant factors in determining validation correctness. Participants exhibited an affirmation bias, causing significantly increased accuracy on valid specifications, but significantly decreased accuracy on invalid specifications. Additionally, participants, particularly those familiar with formal methods, tended to be overconfident in their answers, and be similarly confident regardless of actual correctness.

Our data do not support the belief that formal specifications are inherently human-interpretable to a meaningful degree for system validation. We recommend ergonomic improvements to data presentation and validation training, which should be tested before claims of interpretability make their way back into the formal methods literature.
\end{abstract}

\section{Introduction}

As autonomous systems become increasingly common in daily life, it is critical that their users can be assured of their appropriate behavior. Assurance can be considered along a variety of dimensions, including software verification, statistical testing, and user feedback. One key pillar of assurance in this domain is \textit{interpretability} to humans, 
defined by Kulkarni et al. to be the extent to which an agent's ``behavior [is] consistent with the human’s expectations''
~\cite{interpretable-robotics}.

Formal specifications have recently been presented as one way of making the behavior of autonomous agents interpretable to humans.
Utilizing formal specifications as an output of inference ensures a direct mapping to rules and concepts expressible as natural language \cite{camacho2019learning,li2019formal,decastro2020interpretable,aasi2021inferring}.
Whether these outputs are interpretable is particularly significant given the success of systems that learn specifications from data \cite{bombara2016decision,kong2016temporal,kasenberg2017interpretable,camacho2019learning,aasi2021inferring,decastro2020interpretable,bombara2021offline}. Similarly, formal specifications can tie natural language rules to interpretable plan synthesis, though this is almost always done with human expert translation \cite{kress2009temporal}.

Despite many explicit claims of interpretability, the body of work stops at the automatic construction of plan constraints in formal logic. The translation of such constraints into natural language is then done in a manual and ad hoc manner for the purposes of presenting the associated paper. If we examine some of the actual outputs of these systems, as well as inputs used for plan synthesis (before ad hoc English translations) we find examples such as the following:

A simple case in STL: ``$\phi = (\phi_1 \wedge \phi_2) \vee (\neg \phi_1 \wedge \phi_3), \text{where } \phi_1 = \mathbf{G}_{[470,491]}(z<1.69), \phi_2 = \mathbf{G}_{[473,490]}(v_y \geq 3.31), \text{and } \phi_3 = \mathbf{F}_{[473,491]}(y<10.54)$'' \cite{aasi2021inferring}.

A more complex case encoding some of the rules for a search game in linear temporal logic (LTL):
``$\square(\neg(\bigcirc a^{bone} \wedge \bigcirc a^{yarn})) \wedge \square ((\bigcirc s^{dog} \vee \bigcirc s^{cat} \vee \bigcirc s^{mouse}) \implies (\bigcirc a^{bone} \vee \bigcirc a^{yarn} \vee \bigcirc a^{flute})) \wedge \square (\bigcirc a^{flute} \implies \neg \bigcirc r_{20}) \wedge \square((\bigcirc s^{dog} \wedge (\neg(a^{yarn} \wedge \bigcirc s^{cat}))) \implies \bigcirc a^{bone}) \wedge \bigwedge_{i=1}^{35} \square \Diamond (r_i \vee s^{dog} \vee s^{cat} \vee s^{mouse}) \wedge \square \Diamond (r_1 \vee \neg a^{bone}) \wedge \square \Diamond (r_{32}\vee \neg a^{yarn}) \wedge \square \Diamond (r_{36} \vee \neg a^{flute})$'' \cite{kress2009temporal}.

The latter formula is a subset of the full set of requirements from that paper, where it was split into three parts in the text, which were also provided in natural language.

Such formulas are directly translatable to natural language, and are likely interpretable to some degree by experts in that modeling language and system under study. Still, it is unclear how operationally-relevant claims of interpretability truly are. How well can humans use such formulas to predict model behavior, and which humans are we referring to? Indeed, \textit{explainable} and \textit{interpretable} are ill-defined in the AI/ML literature, and a wide variety of methods are used to measure them, rarely with appeal to human psychology or pedagogical literature, or any evaluation of non-authors' ability to interpret these formulas \cite{miller2017explainable,lipton2018mythos}.

In this work, we address the human-interpretability of formal specifications in the following ways: (Q1) How much of the formal methods literature claiming interpretability actually supports such claims? (Q2) How well can humans validate STL specifications as trajectory constraints? (Q3) How do our findings compare to the literature?
%Q1 is addressed in Section \ref{subsec:claims} of the Background, while Q2 and Q3 are the main focus of this work.

For Q2, the primary focus of this work, we evaluated whether humans are able to use temporal logic specifications to correctly determine if a robot motion plan was valid. Specifically, we ask whether a person's ability to correctly determine whether a given STL specification will generate game-winning motion plans is affected by the presentation of the specification, its complexity, its ground-truth validity, or the demographics of the human interpreter.

This study builds on previous human studies \cite{vinter1998evaluating,loomes1997formal,bollin2014formal,greenman2022little}, but includes participants without experience in formal logic, and focuses on the interpretation of temporal logics as constraints for autonomous systems. Thus, our contributions are more directly applicable to robotics, and particularly to the problem of end-user validation of such systems.

\section{Background}

\subsection{Human-Interpretability of Formal Logic}

There is a strong assumption in the literature that AI systems cast in formal logic are ``human-interpretable'' and/or ``transparent,'' with such claims often making their way into titles and abstracts, and used as justification for formal logic approaches to a variety of problems  \cite{camacho2019learning,li2019formal,decastro2020interpretable,mohammadinejad2020interpretable,aasi2021inferring,bombara2021offline}. 
What the formal logic community seems to mean by ``interpretability'' is that most or all numerical values in the specification have attached semantics, and, sometimes, the specification is not too large or complex, for some (unstated) definitions of large and complex. Such properties might be contrasted with, for example, a deep neural network, which contains a large number of parameters that do not have attached semantics or quickly lose their semantics as one traverses further away from the input and output layers.
% Similarly, a classical AI technique such as a support vector machine (SVM) might also be ``uninterpretable,'' since parameters in these models represent difficult-to-visualize hyperplanes, further obscured from observers by ``kernel tricks,'' making such models only interpretable in very small cases.

Conversely, formal specifications necessarily have formal semantics, and examples from the literature are often compact enough to write in a few paragraphs of text in an academic paper. But do such semantics translate to an appropriate understanding of how specifications constrain plans? Furthermore, how does this extend to the longer, multi-agent, and multi-nested specifications that are required for systems that begin to approach operational complexity?

A small body of work exists in terms of investigations into human reasoning about formal specifications. Vinter et al \cite{vinter1996seven,vinter1998evaluating} evaluated computer scientists trained in the formal specification language Z, finding that these experts did not always reason logically over formal specifications, but rather applied heuristics and biases normally used in natural language, such as a matching bias (seeing cases as relevant in logical reasoning when the lexical content matches that of a provided rule, even when the rule does not necessarily apply). Formal specifications that referred to elements of the real world caused participants to use ``informal'' logical heuristics about the real-world counterpart. %For example, participants tended to (mis)apply Gricean conventions typically used in human social communication to the formal logic situation of statements in Z, leading to errors.
% Furthermore, Vinter et al found that much of the formal logic literature's claims that formal specifications necessarily impute precision and reduce error are not entirely true. 
The strongest predictor of more accurate interpretation was years of experience with formal methods, rather than familiarity with the linguistic construction of the specifications, or the specific type of training received. Short-term coursework was useful for reading specifications, but not for correct interpretation.

In another human study involving students in a ``Specification and Verification'' class, Bollin et al. found that Z specifications were correctly interpreted 89\%, 79\%, and 63\% of the time, for the symbols, logical implications, and incomplete operations in Z \cite{bollin2014formal}. Though this work is presented as a counterpoint to the Vinter et al. studies, it should be noted that this study \emph{only} contained participants in a pedagogical formal methods setting, an even more specific cohort than the Vinter studies. Additionally, the highest performance was shown only in \emph{recognizing symbols} in Z (89\% correct), with the tests that involve more than simple recall showing substantially worse performance (79\%, 63\%).

Finally, Greenman et al. \cite{greenman2022little} conducted a series of studies on the common misconceptions in LTL, considering both English to LTL translations, LTL to English translations, and trace satisfaction evaluations for students in formal methods classes, and researchers who were familiar with LTL. They found a number of common misconceptions (e.g. the semantics of Finally and Until), and errors where Finally and Globally operators were ignored or accidentally introduced. In the translation tasks, many misconceptions occurred in the English to LTL direction, but participants tended to do well in the reverse.

Notably, all the experiments we found on interpretability of formal specifications involved only people who have received at least some training in formal methods.

\subsection{Claims Asserted but Not Supported}
\label{subsec:claims}

Inspired by Miller et al. \cite{miller2017explainable}, we attempted to categorize the research in formal specifications that claim human-interpretability. We collected a set of papers from 2012 to 2022, classified them as \emph{on topic} or \emph{off topic}, and provided a breakdown of the on topic papers according to the evidence they provide in support of interpretability. We performed a search on Web of Science\footnote{\url{https://www.webofscience.com/}} using a query of \texttt{(interpretable OR explainable OR transparency)  ``temporal logic''}, yielding 38 references. Following Miller et al's scoring procedure, the papers were categorized and scored as follows:

\begin{enumerate}
    \item \emph{On topic:} Each paper was categorized as on or off topic, based on whether it contained claims of formal specifications being interpretable by humans (reading/writing specifications, using specifications as an intermediate format within an automated process to (theoretically) improve the interpretability of the process). The one review paper in this set was automatically classified as off-topic.
    
    \item \emph{Data driven:} Each paper was scored from 0 to 2. 1 point was given if one or more of its references was literature from the social sciences validating their method of improving interpretability was appropriate when applied to humans. 2 points were given if the reference criterion was met \emph{and} the referenced method(s) for supporting interpretability were incorporated into an algorithm.
    
    \item \emph{Validation:} Each paper was scored either 0 or 1. 1 point was given if the paper included experiments where human participants (apart from the authors) interpreted the specifications generated by the paper's proposed methodology.
    
\end{enumerate}

Of the 25 on-topic papers, 22 scored a 0, 3 scored a 1, and none scored a 2 on the data-driven criteria. All 25 scored a 0 on validation. This result means that even though 3 of 25 papers supported claims of interpretability with references, \emph{none} incorporated the content of their references into the algorithms they subsequently presented, and \emph{none} tested their interpretability claims with human attempts at interpretation.

On the specification-generation side, Aasi et al presented the only explicit attempt we found to \emph{improve} interpretability in machine-learned specifications, aiming to infer \emph{concise} STL decision trees \cite{aasi2021classification}. %They are able to produce specifications that have a better classification performance and simpler formula structure than previous methods (at the expense of computation time), by combining STL primitives in a tree when possible. 
Like the general claim of formal logic interpretability, this approach appears to be a reasonable heuristic, but the work lacks supporting literature or empirical evidence of efficacy.

\section{Methods}

\subsection{Human Experiments}

We now set out to test claims of formal specification interpretability, focusing on formal logic rules used to describe behaviors of autonomous systems. We conducted an experiment where human participants validated robot motion plans from provided trajectory specifications. 62 adult participants completed the experiment after providing informed consent. The protocol was approved by the MIT Committee on the Use of Humans as Experimental Subjects (protocol E-3748) and the United States Department of Defense Human Research Protection Office (protocol MITL20220002). Participants were recruited from the MIT Behavioral Research Lab participant pool, which includes a diverse set of volunteers from the general public. 
% Only US residency was used as a filter (for compensation reasons).
We also targeted a set of ``expert'' participants via email to formal methods research groups, excluding groups with which the authors were affiliated at the time of writing. Participants were compensated \$5 for completion, plus \$0.67 per correct validation (rounded up), and an additional \$25 for the most correct validations (for a maximum of \$50).

The experiment asked subjects to determine whether a given temporal logic specification would always result in plans that would win a game of capture the flag (CTF) for the blue team, starting from a given game configuration. The survey is broken into three sections:
% \MM{Added transition sentence above and tried to unify the tense (it was mixed past/present. Decided to go with present but we can change that.}

\subsubsection{Demographics}

Participants first provide demographic information focusing on their educational background, particularly around logic, mathematics, natural language processing, and AI. Participants are asked about coursework, experience, and their self-rated familiarity with the topics on five-point Likert questions. Participants can elaborate on their experience with any topic that they did not rate as ``strongly disagree" on the question of familiarity.

\subsubsection{Validation Familiarization}

Participants are introduced to the validation scenarios, with rules for a game of grid-based CTF, explanations of STL notation, and instructions for reading decision trees. Practice questions are provided to check participant understanding of the game mechanics, and of the validation process for each of the three specification formats, along with feedback after each question.

\subsubsection{Validation Testing}

\begin{figure*}
  \centering
  \includegraphics[width=\linewidth]{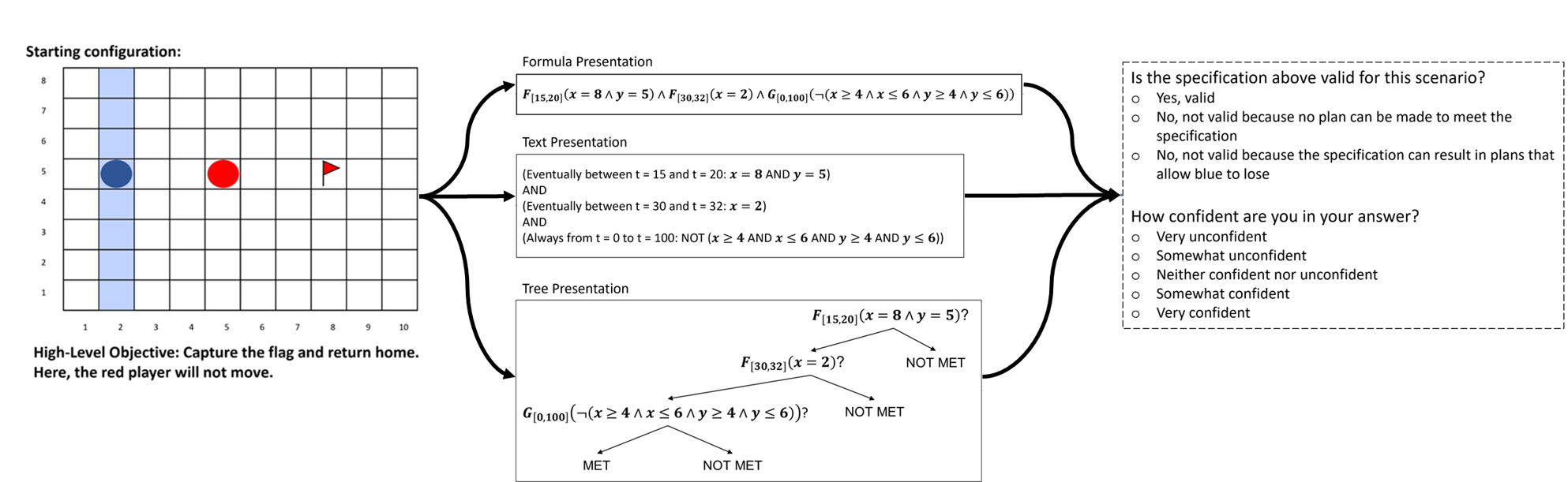}
  \caption{Example validation workflow. For each validation, participants are presented with an initial game configuration and any additional agent behaviors, e.g. red not moving (left), followed by a specification in one of three formats (center), and then two questions about validity and confidence (right).}
  \label{fig:example_scenario}
\end{figure*}

Participants are each presented with 30 random validations out of 60 possible cases, each with a game configuration and a specification for the plans that the blue team can generate (Figure \ref{fig:example_scenario}). Participants are asked about 1) the validity of the configuration/specification pairing (\emph{valid}, \emph{invalid} because no plan can be made to meet the specification, and \emph{invalid} because the specification can result in plans that allow blue to lose) and 2) their confidence in their answer (5-point Likert scale from ``very unconfident'' to ``very confident''). Each validation has an objectively correct answer. Participants must consider both safety and liveness, as blue agents must avoid red agents (safety) and reach a flag and return it to the home region (liveness). Only some scenarios have red agents, and all red agents are explicitly immobile. A specification is valid if and only if \emph{any and all} motion plans satisfying it result in a blue win. %The full introduction to the experiment shown to participants is given in \hl{TODO - Appendix XX}.

% The questions appear in five blocks, with the configuration being the same in each block, and six unique validations being drawn randomly from each block before moving on to the next block. This randomization makes it possible for participants to experience variable numbers of objectively-valid and objectively-invalid validation situations, though the overall number for the entire experiment would be expected to be approximately the same. The randomness was introduced to prevent participants from using process of elimination to help them answer.

% Twelve possible validations are in each block, counterbalanced between the following factors: 1) presentation format (formula, text, and decision tree), and 2) valid and invalid. Each block contains four unique formulas --- two valid, two invalid --- repeated for each presentation format. To control for formula complexity, invalid formulas are a single change away (e.g. single number, operator, or parentheses pair) from one of the valid formulas also in the set. Blocks are presented in a random order.

We hypothesized that the factors of 1) greater participant familiarity with formal methods, 2) higher participant education, 3) formal specifications being presented in a form other than a raw formula, and 4) lower specification complexity, would lead to the responses of 1) increased validation accuracy, and 2) increased validation confidence.

Specification complexity was measured by abstract syntax tree (AST) depth and the number of symbols in the specifications, including duplicates, capturing both the ``nested-ness'' and length of specifications. AST depth varied from 2 to 4, and symbol count varied from 17 to 49 for our specifications.

% Since participants had 30 cases to validate in the main section of the experiment, there was a maximum payout of \$25 before comparing performance across participants. Any ties in top score were decided based on the lowest time spent on the entirety of the validation portion of the experiment.

\subsection{Presentation Formats}

In addition to providing raw STL formulas, two more forms were presented to participants during different validation cases (Figure \ref{fig:example_scenario}, center). The \emph{text}-based presentation translated STL symbols into English text, separated each proposition into its own line, and indented nested blocks. The \emph{tree}-based presentation used the propositions as nodes in a decision tree, matching the output of existing STL inference algorithms  \cite{bombara2016decision,aasi2021inferring}, and the claims in more general machine learning literature that decision trees are an ``intrinsically interpretable'' presentation format for AI models \cite{du2019techniques}. 

\subsection{Demographic Response Coding}

Participants varied in their understanding of the topic area questions in the demographic survey. For example, some participants believed that the use of everyday arithmetic equated to a knowledge of formal logic, while others equated the ability to speak in multiple languages with being familiar with linguistics and natural language processing. To provide a more accurate representation of participant familiarity, we recoded some Likert scores based on participant's open-ended explanations, focusing on formal methods familiarity.
We only decreased coded familiarity in recodes in an attempt to ensure that any claimed topic experts were, in fact, experts. 

% To recode, two experimenters first independently recoded any responses where the participants were judged to have misunderstood the question, and any disagreements were resolved by a third experimenter, who had access to the initial recodes. In total, 9 of the 62 Likert responses for formal methods familiarity were recoded. This was the only demographic question used in our statistical analysis due to potential ambiguity of the other responses. From here on, all references to participant familiarity with formal methods refer to these recoded values, unless otherwise stated.

\subsection{Post-Experiment Questions}

Finally, participants rated their overall validation confidence, ability to understand specifications, and ability to validate specifications under each of the three presentation formats, using 5-point Likert questions. They could also provide free-response explanations for any of those ratings, comments on any features of the specifications that helped or hindered their understanding, as well as thoughts on the experiment as a whole.

\subsection{Statistical Analysis}

We first performed a logistic regression with the predictors of participant age, level of education, (recoded) familiarity with formal methods, presentation format, specification validity, AST depth, and symbol count, and a response variable of validation correctness.
%This regression allowed us to consider the properties of each question, alongside subject-specific properties, which would not be possible with a regression on the subjects' scores.
Next, we performed post-hoc examinations of significant factors from the logistic regression, using Spearman's rank-order correlations for Likert and educational level responses, and dependent t-tests for others, along with a Bonferroni correction for multiple tests, and found Cohen's $d$ effect sizes for significantly different pairs.
% To determine if there were conditions that specifically drove validation correctness in a particular direction, post-hoc examinations were then performed on factors deemed significant. Linear correlation was used in the case of Likert responses, and dependent t-tests were performed for the others, with a Bonferroni correction for multiple tests. Cohen's $d$ effect sizes were calculated for significantly different pairs.
For post-experiment Likert responses, pairwise t-tests with Bonferroni corrections were performed, treating the data as continuous, as recommended by Harpe \cite{harpe2015analyze}.

\section{Results}

\subsection{Objective Validity}

Validation accuracy had a mean and standard deviation of $13.6 \pm 6.1$ out of 30, or $45\% \pm 20\%$. The omnibus logistic regression for validation correctness showed that formal methods familiarity ($p<0.001$), level of education ($p=0.014$), and ground-truth specification validity ($p<0.001$) were significant factors. Formal methods familiarity was significant with both the original and recoded values.

The data did not support specification presentation format, specification complexity, or participant age as significant factors in predicting response correctness (all $p>0.05$).

\begin{figure}
  \includegraphics[width=\linewidth]{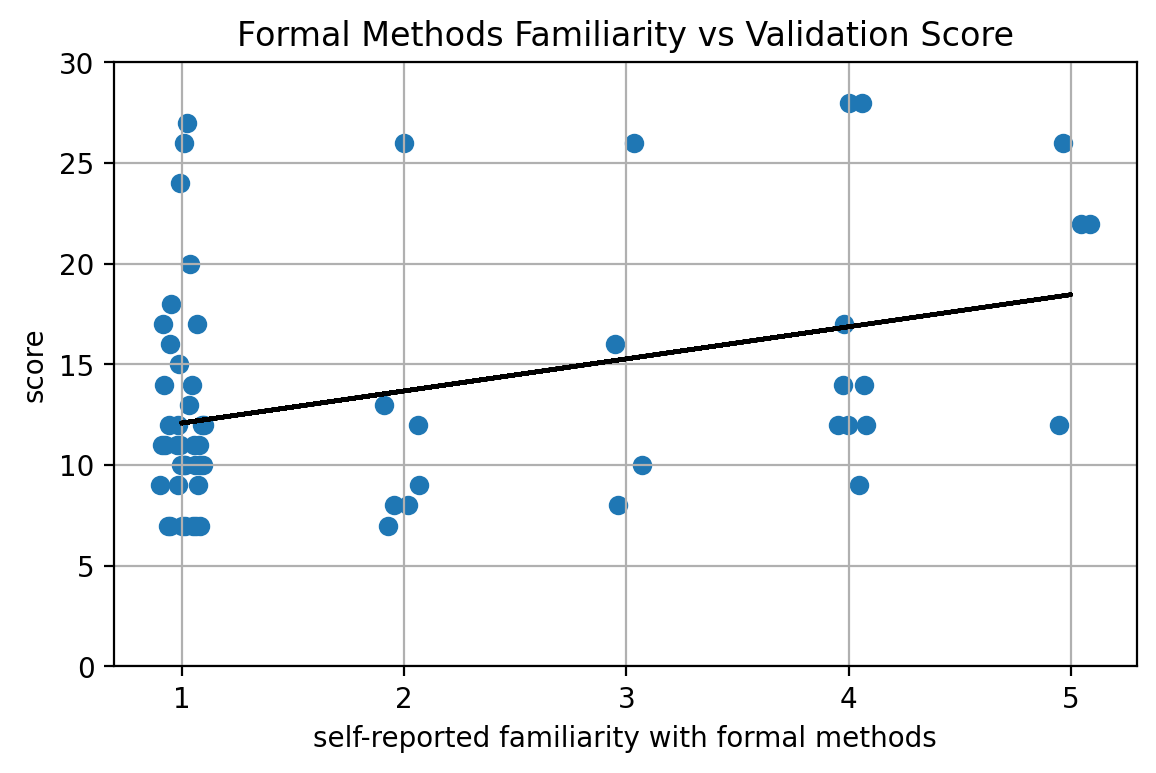}
  \caption{Self-reported familiarity with formal methods vs validation score. Horizontal jitter was added to visually separate points. A linear fit is shown, but correlations were calculated using Spearman's coefficient and were found to be statistically significant ($p=0.019$) and positive ($\rho=0.279$).}
  \label{fig:fm_familiarity_vs_score}
\end{figure}

\begin{figure}
  \includegraphics[width=\linewidth]{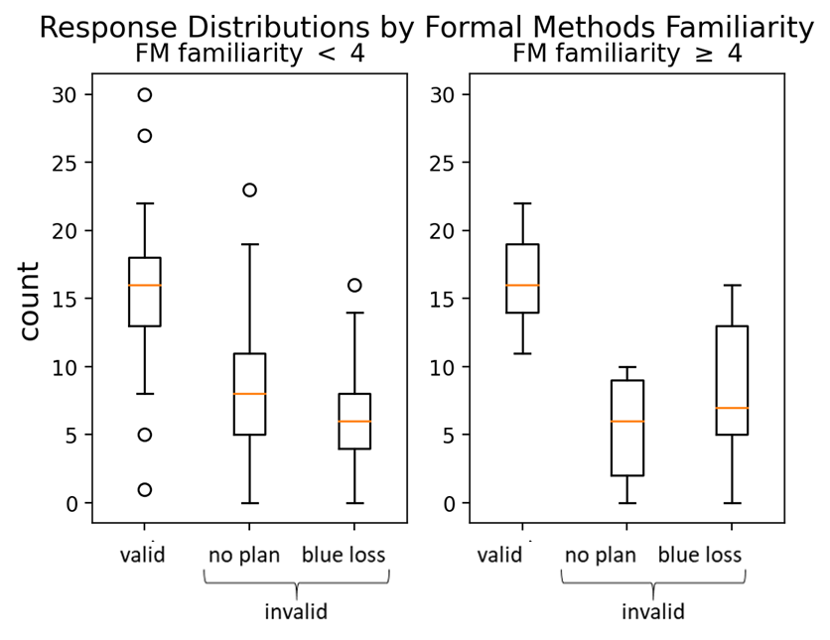}
  \caption{Distribution of participant responses (regardless of correctness) grouped by formal methods expertise ratings. $N = 49$ for FM familiarity $<$ 4, $N = 13$ for FM familiarity $\geq$ 4.}
  \label{fig:response_distribution_by_fm_familiarity}
\end{figure}

Since formal methods familiarity was found to be a significant factor in both the logistic regression and post-hoc rank-order correlation ($p=0.019$, $\rho=0.279$, Figure \ref{fig:fm_familiarity_vs_score}), we conducted a further analysis that grouped participants into ``not-familiar'' (FM familiarity rating $< 4$) and ``familiar'' (rating $\geq 4$) categories and compared whether their base response distributions was the same, regardless of correctness (Figure \ref{fig:response_distribution_by_fm_familiarity}). Independent t-tests showed no significant differences between the groups for any response ($p = 0.025$ for ``invalid - no plan,'' but that is less that the $\alpha = 0.05/3 = 0.017$ Bonferroni threshold). Thus, these results are consistent with the null hypothesis that the response distributions are not different between the two groups.

Although level of education was found to be a significant factor in our analysis under both ordinal and categorical assumptions for variable type, we opted to simply perform a rank-order correlation here (non-significant Spearman correlation at $p=0.079$). Given the variability within each educational category (less than high school, high school, bachelor's, master's, academic/professional doctorate) in terms of actual training received (e.g. STEM vs non-STEM, length, formal methods training, etc), we chose not to examine this factor beyond this coarse evaluation.% Furthermore, raw datapoints are shown rather than boxplots in this case to explicitly show the number of participants in each category.

% \begin{figure}
%   \includegraphics[width=\linewidth]{figures/education_vs_score.png}
%   \caption{Self-reported education level vs validation score. Horizontal jitter was added to visually separate points. Correlation calculations and the regression line were calculated from the non-jittered data. The line is a linear fit, but correlations were calculated using Spearman's rank-order coefficient and were found to be statistically significant ($p=0.045$) and positive ($\rho=0.256$). No participants reported less than high school education.}
%   \label{fig:education_vs_score}
% \end{figure}

\begin{figure}
  \centering
  \includegraphics[width=\linewidth]{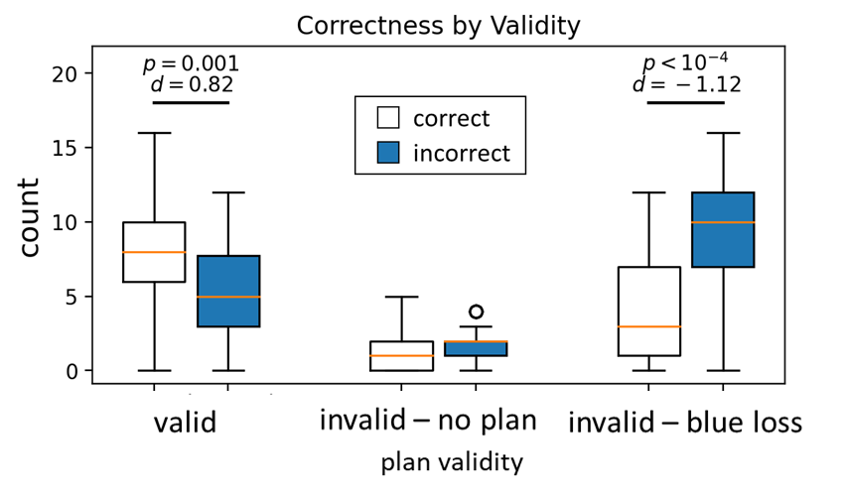}
  \caption{Correct and incorrect validations by ground truth validity. There were an equal number of valid and invalid specifications, but the invalid specifications were split unevenly between no-plan and blue-loss cases.}
  \label{fig:correctness_by_validity}
\end{figure}

Validation accuracy was significantly different depending on the ground-truth validity of specifications ($p < 0.001$, Figure \ref{fig:correctness_by_validity}). Post-hoc t-tests showed that in the cases of valid specifications, participants answered significantly more often correctly than incorrectly with a large effect size ($p=0.001$, $d=0.82$), but in cases where the specification was invalid due to possibility of blue losses, participants were more often incorrect with a large effect size ($p<10^{-4}$, $d=-1.12$).

\subsection{Validation Confidence}

\begin{figure}
  \centering
  \includegraphics[width=\linewidth]{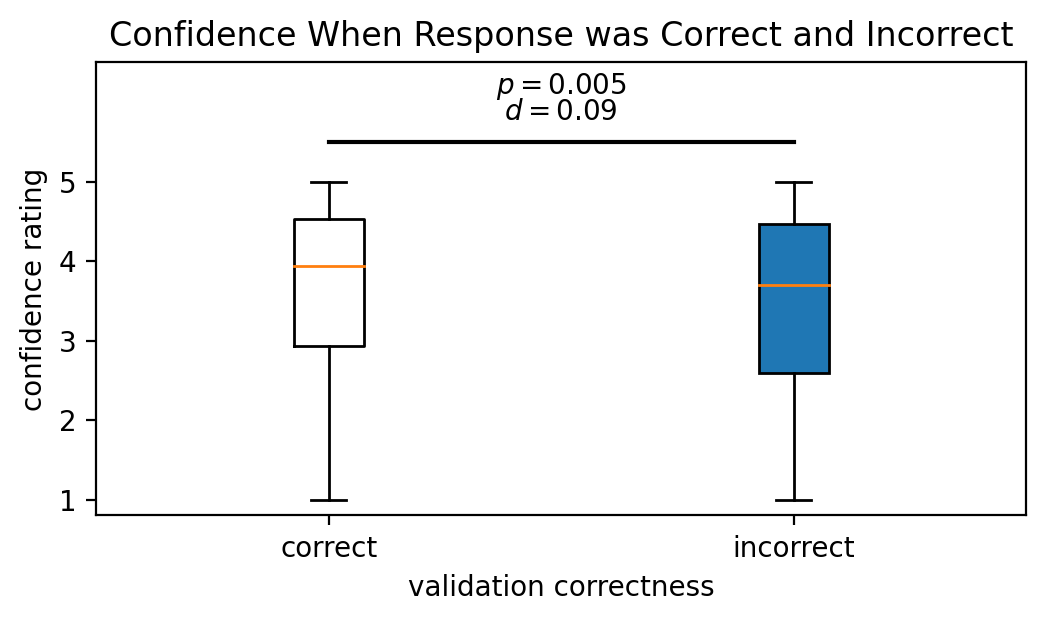}
  \caption{Participant confidence in their answer when their answer was actually correct vs incorrect. Mean confidence was used on a per-participant basis, such that each participant contributed one datapoint to each bar.}
  \label{fig:confidence_vs_correctness}
\end{figure}

Participants' confidence was compared against the actual correctness of their answers, using paired t-tests (Figure \ref{fig:confidence_vs_correctness}). We take the mean confidence of each participant when they were correct and incorrect to construct the box plots, rather than pooling raw confidence values to ensure that each participant contributed one datapoint to each bar. The difference between reported confidence when participants were correct vs incorrect is statistically significant ($p = 0.005$), but the effect size is very small ($d=0.09$), indicating a detectable, but operationally negligible difference.

\begin{figure}
  \includegraphics[width=\linewidth]{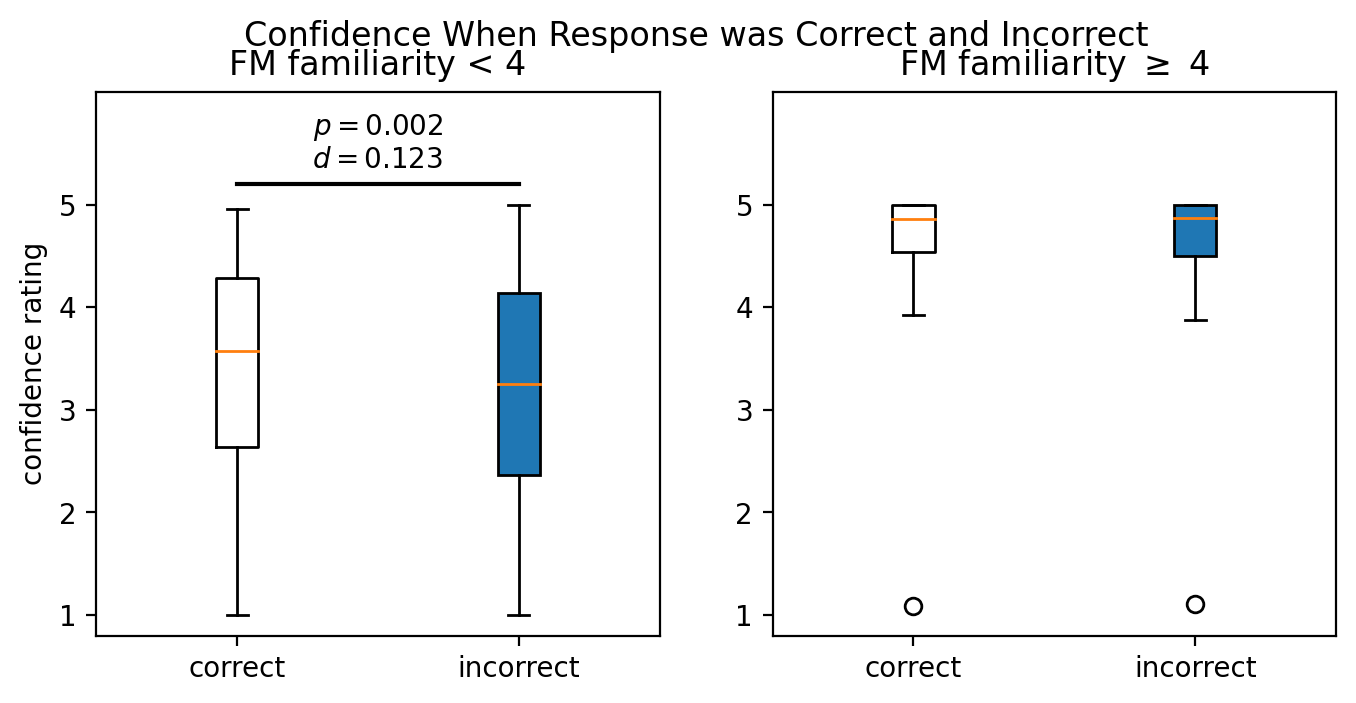}
  \caption{Participant confidence in their answer when their answer was actually correct vs incorrect, split by formal methods familiarity. $N = 49$ for FM familiarity $<$ 4, $N = 13$ for FM familiarity $\geq$ 4.}
  \label{fig:confidence_vs_correctness_split}
\end{figure}

When split into ``familiar'' and ``not-familiar'' formal methods groups (Figure \ref{fig:confidence_vs_correctness_split}), confidence differences between correct and incorrect responses given by unfamiliar participants is statistically significant ($p = 0.002$), but the effect size ($d=0.123$) is again small. Interestingly, among the familiar group, confidence is very high, and is \emph{not} significantly different when the responses were correct or incorrect. These results mean that the data are consistent with a conclusion that members of the ``familiar'' group were equally confident regardless of whether they were correct or not. 

\subsection{Post-Experiment Subjective Evaluations}

\begin{figure*}
  \centering
  \includegraphics[width=0.95\linewidth]{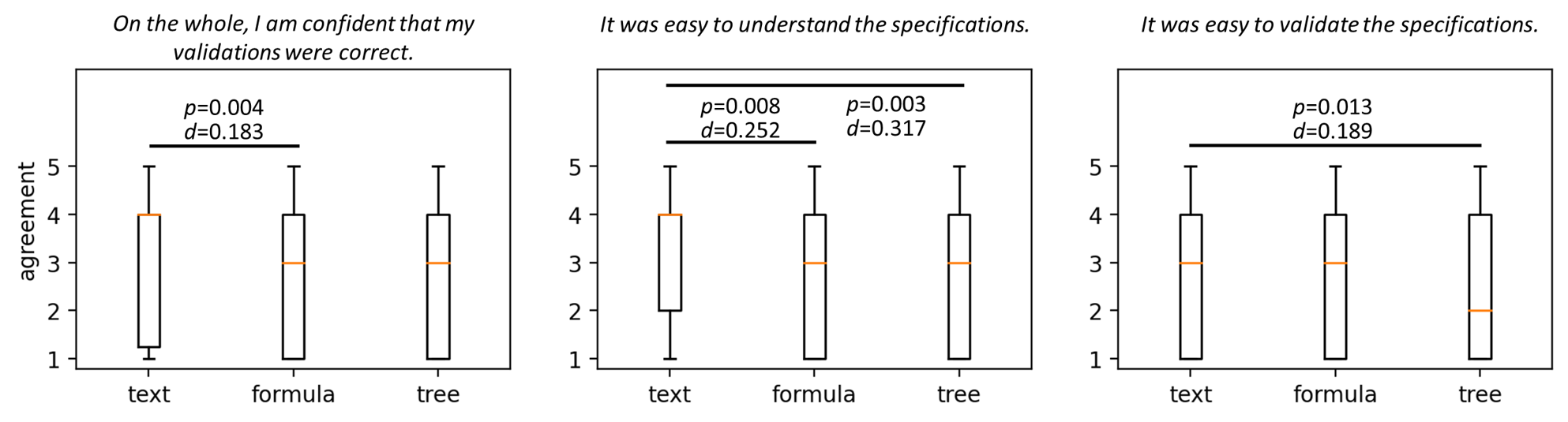}
  \caption{Participant ratings of each presentation format after completing the experiment.}
  \label{fig:post_format_ratings}
\end{figure*}

In post-experiment questions regarding their perceptions of the presentation formats (Figure \ref{fig:post_format_ratings}), participants tended to prefer the text-based presentations over the others. Note that text vs tree on the confidence question had $p = 0.019$ and $d = 0.216$, though this is not considered significant due to the Bonferroni correction. The associated effect sizes are small in all cases, though it is interesting to see these post-experiment preferences despite a lack of significant difference in performance between the three formats. In open-ended responses, multiple participants emphasized the importance of explicitly showing the meaning of symbols (i.e. in text form), and how it was more similar to ``natural language'' than the other forms. Participants were most split in their assessment of the tree format, with some finding it the most difficult, and others finding it the easiest, or at least easier than presenting raw formulas. Final open-ended responses saw a number of participants commenting that the overall experiment was more difficult than expected.

\subsection{LTL Reanalysis}

\begin{figure*}
  \centering
  \includegraphics[width=0.95\linewidth]{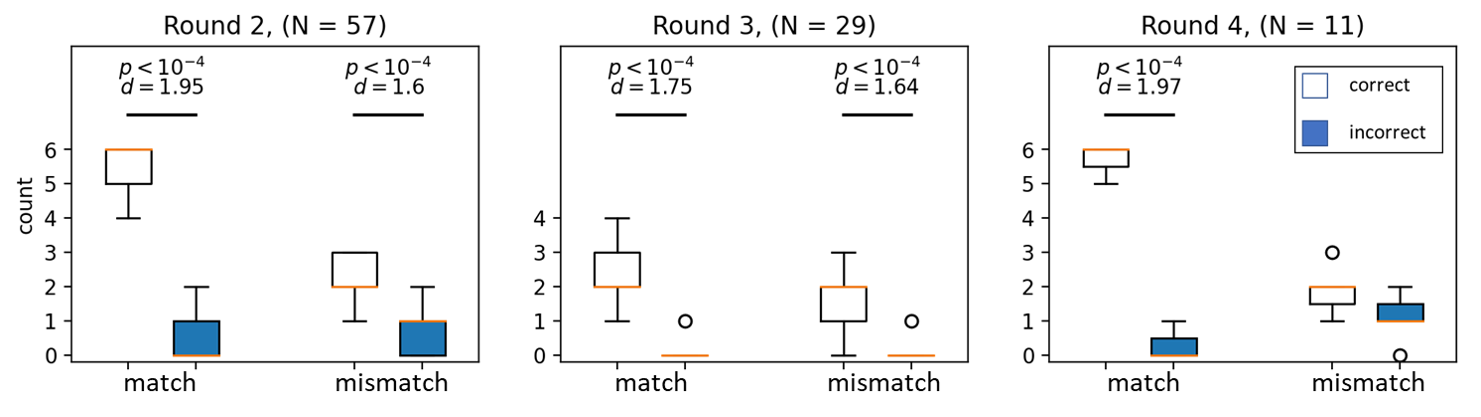}
  \caption{Number of correct and incorrect trace matches by ground-truth validity in data from \cite{greenman2022little}, a study on the interpretability of LTL specifications. Y axis maximums are the maximum possible scores in each round. The terms ``match'' and ``mismatch'' in the trace-matching setting correspond to ``valid'' and ``invalid'' for our experiment. Data from Round 1 was not comparable to our experiment. Details on rounds are in text.}
  \label{fig:ltl_correctness_and_validity}
\end{figure*}

To compare against literature, we reanalyzed the data in Greenman et al. \cite{greenman2022little}, focusing on the portions of the study where participants answered whether a given LTL trace satisfied a given specification (Figure \ref{fig:ltl_correctness_and_validity}). This portion of Greenman's study was the closest we found to our specification validation, but had subjects examining satisfaction of \emph{specific traces} rather than \emph{all possible traces}, as in ours. 

In the experiment, Round 2 involved students in an ``upper-level, tool-based, applied logic course... at a private US university,'' Round 3 involved researchers who had previous experience with LTL, and Round 4 involved students from a later iteration of the applied logic course, but also had them think out loud (Round 1 did not involve trace satisfaction). The recruitment means that not only are participants in this pool already familiar with LTL, but the students are in a course setting that primed them to think about formal logic.

Their participants' specifications/trace matching rate was higher than our specification/validity matching rate, but notably, the only instance where there was \emph{not} a significant difference between the number of correct and incorrect responses was in invalid scenarios (i.e. ``mismatch,'' Figure \ref{fig:ltl_correctness_and_validity}, right). Higher matching rates may be partly explained by the differing experiment setups (2 options in theirs vs 3 in ours) and participant demographics. Additionally, while still significantly different, we see that the invalid (i.e. ``mismatch'') case in Round 2 is the only other distribution with overlapping box plot whiskers (Figure \ref{fig:ltl_correctness_and_validity} left), indicating greater difficulty in correctly identifying invalid specifications. These results parallel the bias that we see in correctness by ground-truth validity (Figure \ref{fig:correctness_by_validity}).

\section{Discussion}

\subsection{Experiment Setup Comparison to Previous Studies}

Several major differences exist between this experiment and previous work \cite{vinter1996seven,vinter1998evaluating,bollin2014formal,greenman2022little} apart from the language used. We do not restrict our participant pool to trained individuals, though we do consider them as a sub-population, a design decision made on the assumption that most users of robotic systems would \emph{not} be expected to have formal methods training. Also, we consider several completely different representations of the same formulas, and treat specification complexity explicitly as a factor.

Finally, while previous work considered state- and trace-satisfaction of formulas (whether a trajectory met a formula), we ask whether \emph{any trajectory meeting the specification guarantees a desired outcome}. We believe that this approach is closer to a practical workflow, as it seems more likely that an operator would check a series of rules that a system is programmed with (or learns) and determine their validity in a given use case, rather than perform numerous state- or trace-matches. Our framing also likely matches the formal methods literature claims more directly insofar as the utility of ``interpretable'' specifications for system safety.

\subsection{Validation Accuracy}

These results strongly disfavor the claim that formal semantics are inherently human-interpretable to a useful degree for system validation. We should note that statistical significance between counts of correct and incorrect responses does not endorse interpretability claims under those conditions even when the direction of the difference is favorable. Significance indicates a detectable difference, a necessary but insufficient condition for useful validation.

It also cannot be claimed that the difficulty of interpreting formal specifications is necessarily due to lack of familiarity with or forgetfulness about the symbols, since the text presentation translated all symbols into English, and a subset of our participants indicated familiarity with formal methods.

While formal methods familiarity correlated with increased score, there was nontrivial variability in scores among ``familiar'' participants (Figure \ref{fig:fm_familiarity_vs_score}). This group also had high confidence regardless of correctness, in contrast to the slight, but detectable confidence difference among ``unfamiliar'' participants (Figure \ref{fig:confidence_vs_correctness_split}). %This result mirrors the sentiment about interpretability in the formal methods literature itself, with experts claiming a certain amount of interpretability in their approaches that may not truly be present.
In fairness, formal methods experts often interact with specifications at a higher level than validation (e.g. using formal specifications for learning system properties), and are not necessarily practiced in manual validation. The latter may require different training, even though it uses the same symbols and logic. It is possible that familiarity with ``breaking of rules'' may be more useful than familiarity with ``construction of rules,'' and training in auditing and quality assurance may be more helpful here.

The bias towards affirming specification validity regardless of actual validity (Figure \ref{fig:correctness_by_validity}) matches the psychological bias towards processing positive and negative information (negative cases take longer to process, and people tend not to want to expend cognitive effort) \cite{wason1959processing}, and the related status quo bias \cite{samuelson1988status}. These biases were also found in the specific application of validating formal specifications in Z \cite{vinter1998evaluating}, and in our re-analysis of the LTL evaluation data \cite{greenman2022little} (Figure \ref{fig:ltl_correctness_and_validity}).

This affirmation bias is particularly concerning for system validation, as failure modes under a given specification are more likely to be overlooked. The particularly high error rate when the specification allows for blue losses (Figure \ref{fig:correctness_by_validity}, right) may point to a satisficing heuristic on the participants' part, where they are only looking for a single winning plan per specification, rather than considering all possible plans \cite{simon1956rational}.

It is interesting that specification complexity was not found to be a significant predictor of response correctness.
However, there is a slight trend towards more incorrect responses for higher values of both complexity measures, perhaps lending some support to the approach taken by Aasi et al. \cite{aasi2021inferring}.
% However, we do see a slight trend towards more incorrect responses in the boxplots for both types of complexity measures (Figures \ref{fig:correctness_by_ast_depth} and \ref{fig:correctness_by_symbol_count}), perhaps lending some amount of support to the approach taken by Aasi et al. \cite{aasi2021inferring}.
Our specifications were relatively simple and succinct in comparison to the rules required for all but the simplest of mobile robot behaviors. Given the trends in the data, it is quite possible that complexity effects are simply masked by other, more dominant factors in this experiment.

\subsection{Applicability of Results to Operational Use Cases}

When considering the real-world applicability of our results, we note several differences between this experiment and operational conditions. Due to these differences, we do not attempt to tie our results directly to the performance of humans in real-world robot validation. Instead, we use them as a starting point to think about what the formal logic community may be missing when claiming interpretability, and how such claims may play out in the real world.

\subsubsection{Training}

Since our validation scenarios were created specifically for this experiment, it is likely fair to treat all participants as being new to this kind of task. As Vinter's experiments show, formal logic training is not as good of a predictor of validation accuracy as years of experience with formal logic \cite{vinter1998evaluating}. A meaningful cohort of robot \textit{users} with experience with formal specifications is unlikely to exist, but this work points to the need for such skills in at least some significant subset of robot users if user-validated autonomy is to be achieved in the way that the formal methods literature suggests. There are many claims to interpretability in the AI literature, but the flip side of training human-interpreters is much less developed and warrants further work.

Participant comments indicated that unfamiliarity with STL symbols was a hindrance, and that there was a strong preference for word-based formats over formulas. Still, comparisons of validation accuracy by format did not confirm this preference as leading to better performance. On the other hand, simply translating formal logic into natural language, and vice-versa, have their own pitfalls \cite{vinter1996seven,vinter1998evaluating,greenman2022little}.

\subsubsection{System Familiarity and Base Rate Validity}

The novelty of this task domain meant that participants did not have a sense of the system's base-rate validity. Real-world users are more likely to have a preconceived notion of such a base rate, though appropriate use of base rates in predicting validity is likely to be superseded by an individual's recall of specific examples \cite{tversky1981evidential}. It remains to be seen how these base rate perceptions interact with the affirmation bias observed here.

\subsubsection{Stakes}

Our subject compensation scaled approximately linearly with correct validations, but this structure may not match that of real-world validation. The latter depends on the setting, but would likely involve more negative incentives --- punishing incorrect validations through system failures (potentially leading to employment, legal, safety, or other impacts), with similarly highly variable incentive magnitudes. Behavioral economics literature on the greater effects of loss aversion over gains \cite{tversky1991loss} suggests that a loss-centric framing may be worth investigating. %One way to do this may be to simply promise a certain amount of compensation upfront, and visibly decrease it (perhaps nonlinearly) for incorrect responses, though such a setup may raise questions of experiment participation and completion.

\subsection{Recommendations and Future Work}

\subsubsection{Claims of Interpretability}

We strongly recommend that the formal methods community presently stop making the broad claim that formal semantics are inherently human-interpretable, or that this is an advantage in operational use cases. Such claims require far more specificity and evidence than the field has customarily provided, without which they are effectively meaningless. Even the more specific assumption (never outright stated) that formal semantics can be used for system validation by people who are trained in their interpretation, is not well-supported by the evidence.

\subsubsection{Design for Interpretability}

However, we do not entirely dismiss the notion that formal semantics may be helpful in system understanding. The psychological basis for interpreting logical statements \cite{wason1959processing} and specific examples \cite{tversky1981evidential} may help in the design of decision aids for human understanding of formal specifications. We propose three areas of exploration to this end: 1) presenting multiple specifications for a user to determine \emph{which} is valid rather than determining whether a \emph{specific} specification is valid, to avoid positive information bias; 2) presenting both plan specifications and associated example executions to ground the former; and 3) organizing/translating specifications into equivalent forms that support human use \cite{servan1990learning}.

\subsubsection{Training for Interpretability}

There is also an open question of human training. Present training in formal methods does not include significant practice in manual specification checking, particularly for cases beyond a very small number of predicates, or beyond matching individual states/traces with specifications. But if humans are to use formal specifications as a means of assurance, different training might be considered. It is possible that there are lessons to be learned from the training required for professions such as quality assurance, auditing, and various legal professions which deal with finding ``rule-breaking'' cases.

% \KL{We might perhaps suggest tailoring specification languages for domain-specific interpretability? I'm thinking something like CaTL for agent counts, but perhaps there is a more general way to ground predicates, signals, etc. that aids in interpretability?}

\section{Conclusions}

Recent work in the formal logic community has often been motivated by a belief that formal specifications are inherently human-interpretable and useful for system validation. Little evidence supports this belief, and the vast majority of the literature does not reference any supporting social science concepts, much less human experiments. This work is the first to our knowledge to consider the problem of validating the safety and liveness of a system constrained by temporal logic specifications, and the first to include participants who were not trained in formal methods before the experiment. 

Current evidence from human testing suggests that temporal logic specifications are \emph{not} inherently human-interpretable. Furthermore, it indicates that the formal methods community, much like our subset of formal methods expert participants, are overconfident in their estimation of interpretability. While it is possible that formal semantics can be helpful for system validation, more work is required to develop the general design and/or human factors principles needed to aid humans in correctly understanding and validating formally-specified system behavior.

\section*{Acknowledgments}

Certain data included herein are derived from Clarivate Web of Science. © Copyright Clarivate 2022. All rights reserved.

\bibliographystyle{IEEEtran}
\bibliography{references.bib}

% Generated by IEEEtran.bst, version: 1.14 (2015/08/26)
\begin{thebibliography}{10}
\providecommand{\url}[1]{#1}
\csname url@samestyle\endcsname
\providecommand{\newblock}{\relax}
\providecommand{\bibinfo}[2]{#2}
\providecommand{\BIBentrySTDinterwordspacing}{\spaceskip=0pt\relax}
\providecommand{\BIBentryALTinterwordstretchfactor}{4}
\providecommand{\BIBentryALTinterwordspacing}{\spaceskip=\fontdimen2\font plus
\BIBentryALTinterwordstretchfactor\fontdimen3\font minus
  \fontdimen4\font\relax}
\providecommand{\BIBforeignlanguage}[2]{{%
\expandafter\ifx\csname l@#1\endcsname\relax
\typeout{** WARNING: IEEEtran.bst: No hyphenation pattern has been}%
\typeout{** loaded for the language `#1'. Using the pattern for}%
\typeout{** the default language instead.}%
\else
\language=\csname l@#1\endcsname
\fi
#2}}
\providecommand{\BIBdecl}{\relax}
\BIBdecl

\bibitem{interpretable-robotics}
A.~Kulkarni, S.~Sreedharan, S.~Keren, T.~Chakraborti, D.~E. Smith, and
  S.~Kambhampati, ``Designing environments conducive to interpretable robot
  behavior,'' in \emph{2020 IEEE/RSJ International Conference on Intelligent
  Robots and Systems (IROS)}, 2020, pp. 10\,982--10\,989.

\bibitem{camacho2019learning}
A.~Camacho and S.~A. McIlraith, ``Learning interpretable models expressed in
  linear temporal logic,'' in \emph{Proceedings of the International Conference
  on Automated Planning and Scheduling}, vol.~29, 2019, pp. 621--630.

\bibitem{li2019formal}
X.~Li, Z.~Serlin, G.~Yang, and C.~Belta, ``A formal methods approach to
  interpretable reinforcement learning for robotic planning,'' \emph{Science
  Robotics}, vol.~4, no.~37, p. eaay6276, 2019.

\bibitem{decastro2020interpretable}
J.~DeCastro, K.~Leung, N.~Ar{\'e}chiga, and M.~Pavone, ``Interpretable policies
  from formally-specified temporal properties,'' in \emph{2020 IEEE 23rd
  International Conference on Intelligent Transportation Systems (ITSC)}.\hskip
  1em plus 0.5em minus 0.4em\relax IEEE, 2020, pp. 1--7.

\bibitem{aasi2021inferring}
E.~Aasi, C.~I. Vasile, M.~Bahreinian, and C.~Belta, ``Inferring temporal logic
  properties from data using boosted decision trees,'' \emph{arXiv preprint
  arXiv:2105.11508}, 2021.

\bibitem{bombara2016decision}
G.~Bombara, C.-I. Vasile, F.~Penedo, H.~Yasuoka, and C.~Belta, ``A decision
  tree approach to data classification using signal temporal logic,'' in
  \emph{Proceedings of the 19th International Conference on Hybrid Systems:
  Computation and Control}, 2016, pp. 1--10.

\bibitem{kong2016temporal}
Z.~Kong, A.~Jones, and C.~Belta, ``Temporal logics for learning and detection
  of anomalous behavior,'' \emph{IEEE Transactions on Automatic Control},
  vol.~62, no.~3, pp. 1210--1222, 2016.

\bibitem{kasenberg2017interpretable}
D.~Kasenberg and M.~Scheutz, ``Interpretable apprenticeship learning with
  temporal logic specifications,'' in \emph{2017 IEEE 56th Annual Conference on
  Decision and Control (CDC)}.\hskip 1em plus 0.5em minus 0.4em\relax IEEE,
  2017, pp. 4914--4921.

\bibitem{bombara2021offline}
G.~Bombara and C.~Belta, ``Offline and online learning of signal temporal logic
  formulae using decision trees,'' \emph{ACM Transactions on Cyber-Physical
  Systems}, vol.~5, no.~3, pp. 1--23, 2021.

\bibitem{kress2009temporal}
H.~Kress-Gazit, G.~E. Fainekos, and G.~J. Pappas, ``Temporal-logic-based
  reactive mission and motion planning,'' \emph{IEEE transactions on robotics},
  vol.~25, no.~6, pp. 1370--1381, 2009.

\bibitem{miller2017explainable}
T.~Miller, P.~Howe, and L.~Sonenberg, ``Explainable ai: Beware of inmates
  running the asylum or: How i learnt to stop worrying and love the social and
  behavioural sciences,'' \emph{arXiv preprint arXiv:1712.00547}, 2017.

\bibitem{lipton2018mythos}
Z.~C. Lipton, ``The mythos of model interpretability: In machine learning, the
  concept of interpretability is both important and slippery.'' \emph{Queue},
  vol.~16, no.~3, pp. 31--57, 2018.

\bibitem{vinter1998evaluating}
R.~Vinter, ``Evaluating formal specifications: a cognitive approach,'' 1998.

\bibitem{loomes1997formal}
M.~Loomes and R.~Vinter, ``Formal methods: No cure for faulty reasoning,'' in
  \emph{Safer Systems}.\hskip 1em plus 0.5em minus 0.4em\relax Springer, 1997,
  pp. 67--78.

\bibitem{bollin2014formal}
A.~Bollin and D.~Rauner-Reithmayer, ``Formal specification comprehension: the
  art of reading and writing z,'' in \emph{Proceedings of the 2nd FME Workshop
  on Formal Methods in Software Engineering}, 2014, pp. 3--9.

\bibitem{greenman2022little}
B.~Greenman, S.~Saarinen, T.~Nelson, and S.~Krishnamurthi, ``Little tricky
  logic: Misconceptions in the understanding of {LTL},'' \emph{The Art,
  Science, and Engineering of Programming}, vol.~7, 2023.

\bibitem{mohammadinejad2020interpretable}
S.~Mohammadinejad, J.~V. Deshmukh, A.~G. Puranic, M.~Vazquez-Chanlatte, and
  A.~Donz{\'e}, ``Interpretable classification of time-series data using
  efficient enumerative techniques,'' in \emph{Proceedings of the 23rd
  International Conference on Hybrid Systems: Computation and Control}, 2020,
  pp. 1--10.

\bibitem{vinter1996seven}
R.~Vinter, M.~Loomes, and D.~Kornbrot, ``Seven lesser known myths of formal
  methods: uncovering the psychology of formal specification,'' 1996.

\bibitem{aasi2021classification}
E.~Aasi, C.~I. Vasile, M.~Bahreinian, and C.~Belta, ``Classification of
  time-series data using boosted decision trees,'' \emph{arXiv preprint
  arXiv:2110.00581}, 2021.

\bibitem{du2019techniques}
M.~Du, N.~Liu, and X.~Hu, ``Techniques for interpretable machine learning,''
  \emph{Communications of the ACM}, vol.~63, no.~1, pp. 68--77, 2019.

\bibitem{harpe2015analyze}
S.~E. Harpe, ``How to analyze likert and other rating scale data,''
  \emph{Currents in pharmacy teaching and learning}, vol.~7, no.~6, pp.
  836--850, 2015.

\bibitem{wason1959processing}
P.~C. Wason, ``The processing of positive and negative information,''
  \emph{Quarterly Journal of Experimental Psychology}, vol.~11, no.~2, pp.
  92--107, 1959.

\bibitem{samuelson1988status}
W.~Samuelson and R.~Zeckhauser, ``Status quo bias in decision making,''
  \emph{Journal of risk and uncertainty}, vol.~1, no.~1, pp. 7--59, 1988.

\bibitem{simon1956rational}
H.~A. Simon, ``Rational choice and the structure of the environment.''
  \emph{Psychological review}, vol.~63, no.~2, p. 129, 1956.

\bibitem{tversky1981evidential}
A.~Tversky and D.~Kahneman, ``Evidential impact of base rates,'' Stanford Univ
  Ca Dept Of Psychology, Tech. Rep., 1981.

\bibitem{tversky1991loss}
------, ``Loss aversion in riskless choice: A reference-dependent model,''
  \emph{The quarterly journal of economics}, vol. 106, no.~4, pp. 1039--1061,
  1991.

\bibitem{servan1990learning}
E.~Servan-Schreiber and J.~R. Anderson, ``Learning artificial grammars with
  competitive chunking.'' \emph{Journal of Experimental Psychology: Learning,
  Memory, and Cognition}, vol.~16, no.~4, p. 592, 1990.

\end{thebibliography}

\end{document}